\documentclass[runningheads]{llncs}
\usepackage{graphicx}
\usepackage{multirow}
\usepackage{pgfplots}
\usepackage{bbding}
\usepackage[backref]{hyperref}
\usepackage[numbers,sort&compress]{natbib}
\begin{document}
\title{Full Resolution Repetition Counting}
%
%
\author{Jianing Li \and
Bowen Chen \and
Zhiyong Wang\thanks{Corresponding Author.} \and Honghai Liu}
%
%
\institute{School of Mechanical Engineering and Automation, Harbin Institute of Technology, Shenzhen \\
State Key Laboratory of Robotics and System, Harbin Institute of Technology
\email{yzwang$\_$sjtu@sjtu.edu.cn} \\}
\maketitle            
%
\begin{abstract}
Given an untrimmed video, repetitive actions counting aims to estimate the number of repetitions of class-agnostic actions. To handle the various length of videos and repetitive actions, also optimization challenges in end-to-end video model training, down-sampling is commonly utilized in recent state-of-the-art methods, leading to ignorance of several repetitive samples. In this paper, we attempt to understand repetitive actions from a full temporal resolution view, by combining offline feature extraction and temporal convolution networks. The former step enables us to train repetition counting network without down-sampling while preserving all repetition regardless of the video length and action frequency, and the later network models all frames in a flexible and dynamically expanding temporal receptive field to retrieve all repetitions with a global aspect. We experimentally demonstrate that our method achieves better or comparable performance in three public datasets, i.e., TransRAC, UCFRep and QUVA. We expect this work will encourage our community to think about the importance of full temporal resolution.

\end{abstract}
\section{Introduction}
\begin{figure}[t]
\centering
\includegraphics[width=12cm]{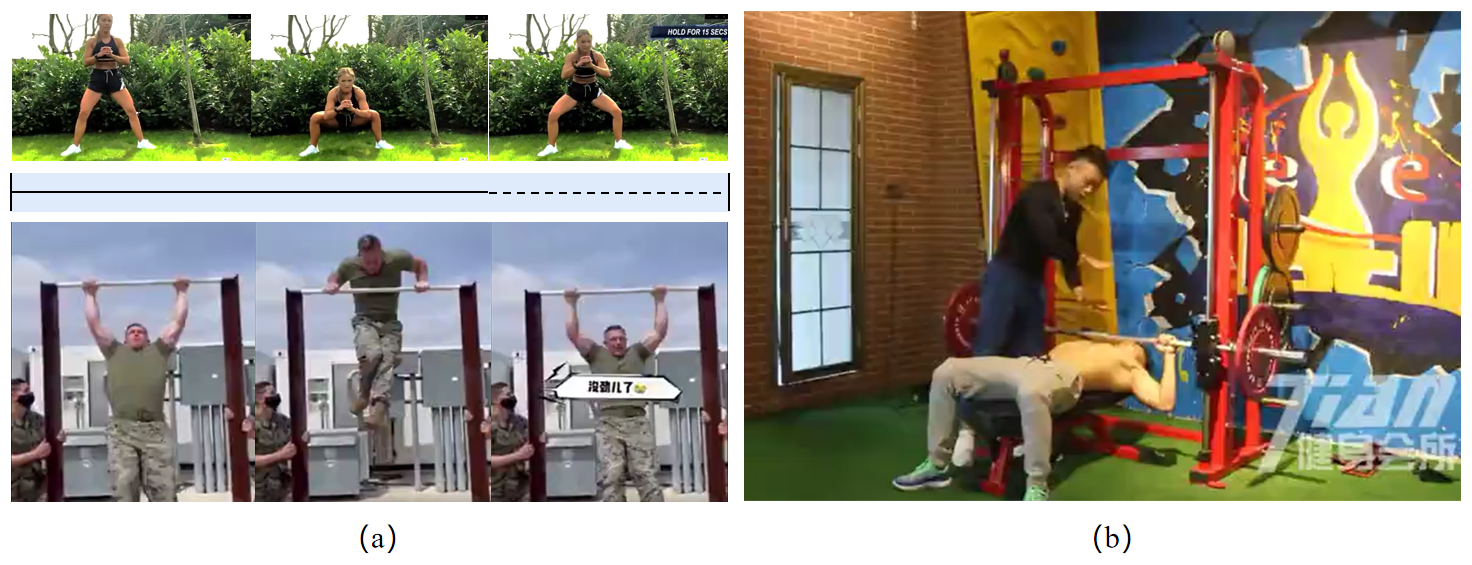}
%
\caption{\label{example} Several challenging examples in RepCount dataset: (a) There are incomplete actions in the videos, which may be mistaken for a repetitive action by the models; (b) There are two people in the video, which may affect the predicted count.} 
\end{figure}

Repetition counting, aiming to count the repetitions of class-agnostic actions, is a fundamental problem in computer vision. It has great importance for analyzing human activities which are commonly involves repetitive actions, such as physical exercise movements. This task is challenging since following challenges: a) various duration of actions within the videos; b) breaks exiting in actions; c)incomplete actions being counted (see Fig. \ref{example}); d) noise in the datasets such as changes in view point, multiple people in videos and so on (see Fig. \ref{example}). These challenges make most of models with down-sampling not perform very well. In terms of down-sampling, earlier works for repetition counting can be grouped into two categories: sliding window \cite{dwibedi2020counting} and down-sampling to fixed frames \cite{levy2015live, zhang2020context, dwibedi2020counting, zhang2021repetitive}. It's hard to choose one optimal window size which is also fixed and sliding window is unable to handle various duration, also leading to context loss. As for down-sampling to fixed frames, too few selected frames may ignore some repetitions, while too much frames will cause computational burden. Recent approaches \cite{hu2022transrac} rely on multi-scale temporal correlation encoder to make up for missing information caused by down-sampling. Despite the success of the multi-scale model, these approaches operate on low temporal resolution of a few frames per second.

In this paper, we introduce a new model which combines offline feature extraction and temporal convolutional networks. In contrast to previous approaches, the proposed model operates on the full temporal resolution and brings performance improvements compared with recent state-of-the-art methods. \textbf{First,} we utilize the full temporal resolution and offline feature extraction for the input videos, which can offer more fine-grained information. \textbf{Second,} how to extract high-level features for long videos is of great importance. Inspired by action segmentation \cite{lea2017temporal, farha2019ms}, we use TCNs as the encoder of the model, which consists of several layers of dilated 1D convolution. The use of dilated convolution enables the model to have a huge temporal receptive field, which can deal well with various duration of actions whether in inter-videos or intra-videos. To the best of our knowledge, we are the first to introduce full temporal resolution into the repetition counting filed. 

In a nutshell, our \textbf{contributions} are three-fold:

1) We first adopt two-phase strategy to understand repetition actions based on offline frame-wise feature extraction. It enables the model to explore extensive temporal context and extract complete motion patterns, which is important for retrieving all repetitions in videos.

2) Temporal convolutional networks is designed to extract high-level features from a global view. We utilize dilated 1D convolution to obtain a huge temporal receptive field, which can capture long-range dependencies as well as inconsistent action periods.

3) Extensive experimental results demonstrate that our method achieves better or comparable performance on three public datasets.

\section{Related Works}
\subsection{\textbf{Repetition Counting}}
Crowd counting and objects counting in images are active fields in computer vision, while repetitive actions counting did not receive much attention. In terms of methods, earlier works is focusing on how to convert the motion field into one-dimensional signals, where peak detection \cite{thangali2005periodic}, Fourier analysis \cite{azy2008segmentation, cutler2000robust, pogalin2008visual, briassouli2007extraction, tsai1994cyclic} can be used. However, they are only suitable for stationary situations. Then some methods pay attention on estimation of action periods.
\cite{levy2015live} uses CNNs to classify cycle lengths within a fixed number of frames, while it does not take complex situations into account, such as variations in video length. Recent approaches \cite{dwibedi2020counting, zhang2020context} propose some novel frameworks for repetition counting. \cite{zhang2020context} propose a context-aware and scale-insensitive framework, which can estimate and adjust the cycle lengths in a coarse-to-fine manner, integrating with a context-aware network. However, it predicts the number of repetitions for down-sampled inputs, and then estimates the count for the entire videos, which does not consider the interruptions or inconsistent action cycles existing in some videos. \cite{dwibedi2020counting} focuses on repetition counting and periodicity detection, which converts the repetition counting task into a per-frame binary classification problem. However, the input of model is consecutive non-overlapping windows of fixed-length frames, which is easy to lose context information if the action cycles are too long. Moreover, there is an upper limit to the predicted period lengths, whose applications are limited. Recently,
\cite{zhang2021repetitive} utilizes the sound for the first time and achieves cross-modal temporal interaction. Though this method also down-samples the videos, it adds a temporal stride decision module to select the best temporal stride for each video. \cite{hu2022transrac} proposes a multi-scale model, which can compensate for the loss of information caused by down-sampling. Whether high and low-frequency actions nor long and short videos, multi-scale fusion can all perform well. The latest method is PoseRAC \cite{yao2023poserac} which is the first pose-level model and outperforms all existing video-level methods.
\subsection{Temporal Convolutional Networks} Temporal Convolutional Networks is a class of time-series models, which contains a set of convolutional filters. \cite{lea2017temporal} introduces two types of TCNs which are Encoder-Decoder TCN and Dilated TCN, whose input and output share the same length. They both use a hierarchy of temporal convolutional filters. ED-TCN consists of a series of long convolutional filters, pooling and upsampling, while Dialted TCN uses a deep stack of dilated convolution with dilated factor. Due to their huge receptive field, they can perform fine-grained action detection and capture action cycles with long-range dependencies.
\subsection{Temporal Self-Similarity Matrix} In counting task, the most important is to explore similar patterns between instances, thus it is crucial to introduce self-similarity matrix into repetition counting task. The most common method to represent the correlation between vectors is dot product. Of course, cosine similarity can also be used. For repetition counting, \cite{dwibedi2020counting} uses the negative of squared euclidean distance as the similarity between frames, and \cite{hu2022transrac} utilizes the attention mechanism \cite{vaswani2017attention} to calculate the similarity, where the attention function can be described as mapping a query and a set of key-value pairs to an output, which can focus on important information in long sequences with less parameters and fast speed.

\begin{figure*}[ht]
\centering
\includegraphics[width=11cm]{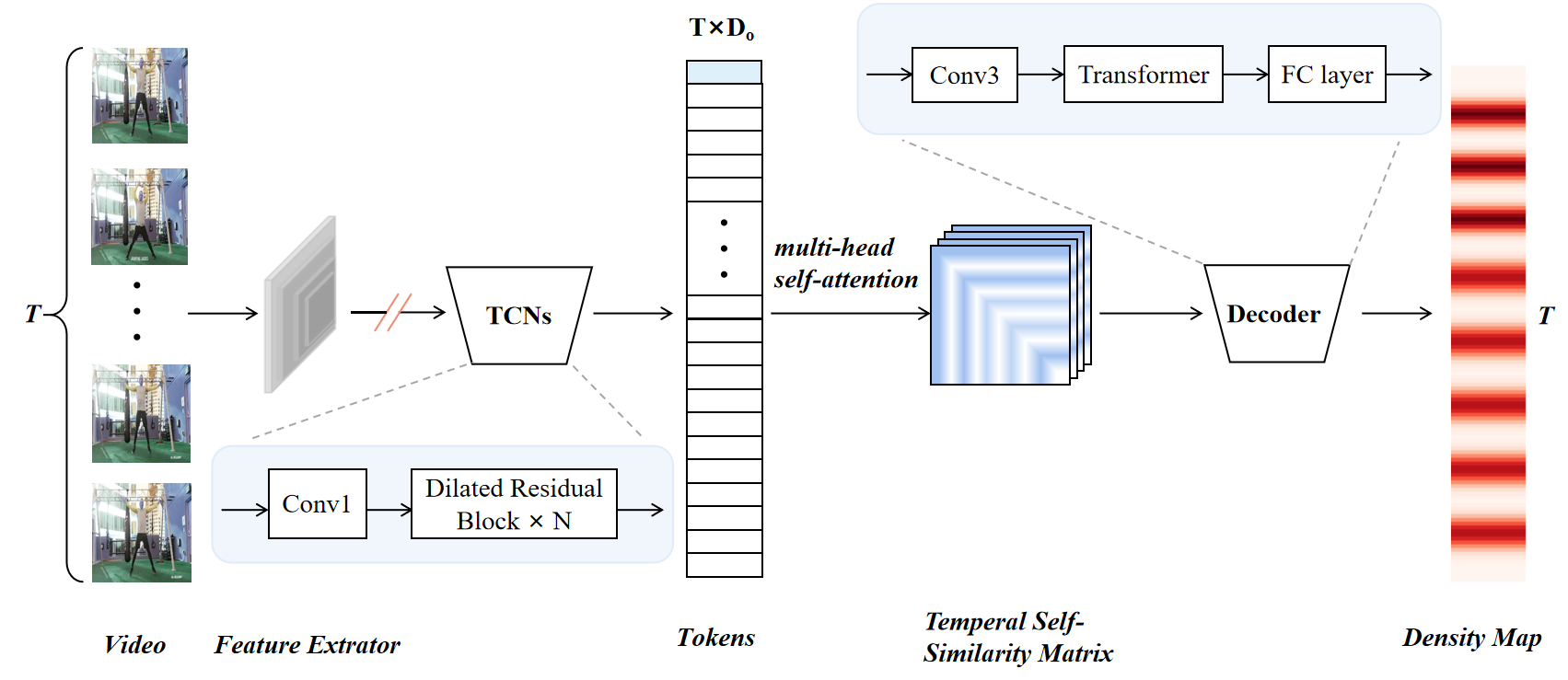}
%
\caption{\label{model}Overview of our proposed model. For an input video, we use the encoder to extract high-level features. Then calculate the similarity between frames and apply the decoder on the temporal self-similarity matrix, which outputs the predicted density map.} 
\end{figure*}

\section{Methodology}
Given a video that contains class-agnostic repetitive actions, our goal is to estimate the number of repetitions. To achieve this, we propose a model based on the full temporal resolution. An overview of our model is depicted in Figure ~\ref{model}. The model is composed of three modules: (1)\textbf{an encoder} which consists of video feature extractor and temporal convolution networks(TCNs), (2)\textbf{temporal self-similarity matrix} which represents the similarity between frames, (3)\textbf{a decoder} which outputs the predicted density map. In the following sections, we present the details of each component. 

\subsection{Encoder}
Our encoder is composed of two main components: video feature extractor and TCNs. Assume that the given video has T frames $F=[f_{1},f_{2},...,f_{t}]$. We extract the features and feed them into the TCNs to produce embeddings $X=[X_{1},X_{2},...,X_{t}]$.

\subsubsection{Video feature extractor.} Processing long videos in both spatial and temporal dimensions is challenging. There are many existing methods such as C3D, SlowFast and I3D which can be used to extract features. Specially, we use a video swin transformer backbone to extract features. Video swin transformer \cite{liu2022video} has several stages which consist of video swin transformer block and patch merging. It performs well in both effect and efficiency. RGB frames are fed as an input clip to the video swin transformer network and the output is of size $7 \times 7 \times D_{0}$. We apply a layer of global average pooling to get the final tokens $1 \times D_{0}$. All the tokens (outputs of video swin transformer) are stacked along the temporal dimension and thus form a $T \times D_{0}$ video token representation, which is the input of TCNs.

\subsubsection{Temporal Convolution networks (TCNs).} TCNs uses a deep stack of dilated convolutions to capture long-range dependencies. Compared with vanilla convolution, dilated convolution has a dilation rate parameter, which indicates the size of the expansion. Without increasing the number of parameters, dilated convolution has a huge receptive field.

\begin{figure}[h]
\centering
\includegraphics[width=4cm]{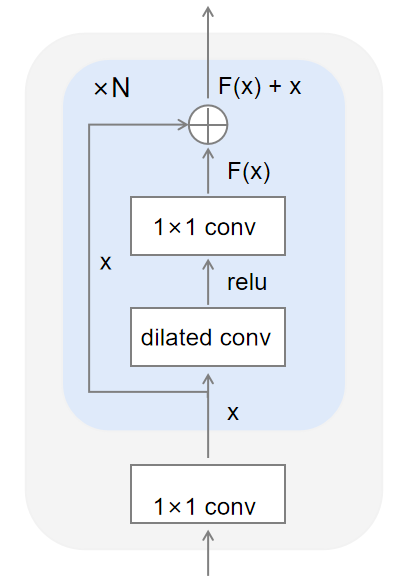}
%
\caption{\label{TCNs}Overview of the TCNs. TCNs is composed of one $1 \times 1$ convolution layer and several dilated residual blocks.} 
\end{figure}

TCNs consists of a $1 \times 1$ convolution layer which can adjust the dimension of tokens, and $\times N$ dilated residual blocks (see Figure ~\ref{TCNs}). Each block has the same structure, which contains a dilated convolution layer, ReLU activation and $1 \times 1$ convolution layer. As the number of block N increases, the dilation factor is doubled. However, there are also some problems with the dilated convolution, such as gridding problem, which may lead to some tokens that are underutilized. Despite larger receptive field to capture long-range temporal patterns, some long-distance information that are completely uncorrelated with current frame will affect the consistency of the data, which is detrimental to shorter action cycles. Thus we further add skip connections. The output of each block is added to the output of the previous block as the input of the next block. 

\subsection{Temporal Self-Similarity Matrix} We use multi-head attention \cite{vaswani2017attention} to generate temporal self-similarity matrix. The input contains queries($Q$) and keys($K$) of dimension $D_{k}$, and values($V$) of dimension $D_{v}$. Specially, queries, keys and values are the same in our model, which are the output of TCNs. Then we compute the dot product of the queries with all keys and divide each by $\sqrt{D_{k}}$ with a softmax function to get the attention scores, which form the temporal self-similarity matrix whose size is $T \times T \times heads$. The similarity matrix can roughly present the distribution of repetitive actions.
The set of operations can be formally described as follows:
\begin{equation}
    score = softmax(\frac{QK^{T}}{\sqrt{D_{k}}})
\end{equation}

\subsection{Decoder} We apply a 2D convolution and a fully connected layer on similarity matrix. After that, similarity matrix needs to be flattened into the sequence whose size is $T \times D$. Following this, the sequence with position embeddings is fed as an input into the transformer \cite{vaswani2017attention}. Finally, the output passes through the fully connected layer to get the predicted density map $D$, which is of size $T \times 1$. 

The ground-truth density map $D^{gt}$ is generated by Gaussian functions, where each repetitive action corresponds to a Gaussian distribution and the mean of the Gaussian function is in the mid-frame. The density map indicates contribution of each frame to the complete action. Thus the count is the sum of the density map:
\begin{equation}
    c = \sum_{i=1}^T D_{i}
\end{equation}
where $D_{i}$ is the predicted value of each frame.

We use the Euclidean distance between the predicted density map $D$ and ground truth $D^{gt}$ as the loss function, which is defined as follows:
\begin{equation}
    L = {\Vert D - D^{gt}\Vert}^{2}_{2}
\end{equation}

\section{Experiment}
\subsection{Experiment Setup}
\subsubsection{Datasets.} Our experiments are conducted on three datasets: RepCount\cite{hu2022transrac}, QUVA\cite{dwibedi2020counting}, UCFRep\cite{zhang2020context}. The \textbf{RepCount} dataset provides fine-grained annotations in the form of start and end of actions, while two other datasets only provide the start or end of actions. In RepCount dataset, there are 758 videos used for training, 132 videos for validation and 151 videos for testing. RepCount dataset covers a large number of video length variations and contains anomaly cases, thus is more challenging than other datasets. The \textbf{QUVA} dataset is composed of 100 videos for testing with a wide range of repetitions, where each video contains 12.5 action cycles in average. It includes action videos in realistic scenarios with occlusion, changes in view point, and inconsistency in action cycles. The \textbf{UCFRep} dataset contains 526 repetitive action videos, which are collected from the dataset UCF101 \cite{soomro2012ucf101}. The original UCF101 is an action recognition dataset collected from YouTube, which can be classified into 101 action categories. The details about datasets is shown in Tab.~\ref{table_dataset}.

\begin{table*}[h]
\centering\setlength\tabcolsep{5pt}
\caption{Dataset statistic of RepCount, UCFRep and QUVA.}
\label{table1}
\begin{tabular}{c|c|c|c}
\hline
\multicolumn{1}{c|}{} & \multicolumn{1}{|c|}{RepCount} & \multicolumn{1}{|c|}{UCFRep} & \multicolumn{1}{|c}{QUVA} \\
\hline

\multicolumn{1}{c|}{Num. of Videos} & 1041 & 526 & 100 \\

\multicolumn{1}{c|}{Duration Avg. $\pm$ Std.} & 30.67 $\pm$ 17.54 & 8.15 $\pm$ 4.29 & 17.6 $\pm$ 13.3 \\

\multicolumn{1}{c|}{Duration Min. $\pm$ Max} & $4.0 / 88.0$ & $2.08 / 33.84$ & $2.5 / 64.2$\\

\multicolumn{1}{c|}{Count Avg. $\pm$ Std.} & 14.99 $\pm$ 17.54 & 6.66 & 12.5 $\pm$ 10.4 \\

\multicolumn{1}{c|}{Count Min. $\pm$ Max} & $1 / 141$ & $3 / 54$ & $4 / 63$\\
\hline
\end{tabular}
\label{table_dataset}
\end{table*}

\subsubsection{Evaluation Metric.} Following the previous work, we use Mean Absolute Error(MAE) and Off-By-One(OBO) count errors to evaluate the proposed method. MAE and OBO are defined as follows:
\begin{equation}
    OBO = \frac 1N \sum_{i=1}^N [|\widetilde{c_i}-c_i|\le1]
\end{equation}
\begin{equation}
    MAE = \frac 1N \sum_{i=1}^N \frac{|\widetilde{c_i}-c_i|}{\widetilde{c_i}}
\end{equation}

where $\widetilde{c}$ is the ground truth repetition count and $c$ is the predicted count. N is the number of given videos. 

\subsubsection{Implementation Details.} In the proposed network, we use the output features of video swin transformer (after global average pool) as the inputs of TCNs, and thus $D_{0}=768$. Taking the computational cost and performance into account, the frame rate is set to 5. Due to various length of videos, we need to pad feature vectors to the same length. In TCNs, the number of dilated residual blocks is set to 6. In the process of calculating the similarity matrix, the dimension of queries, keys and values is 512. Our model is implemented in PyTorch and trained on two NVIDIA GeForce RTX 3090 GPUs. We train the model for 200 epoches with a learning rate of $ 8\times10^{-6}$ with Adam optimizer and batch size of 48 videos. Testing is conducted on the same machine.

\subsection{Evaluation and Comparison} We compare our model with existing video-level methods on RepCount, UCFRep and QUVA datasets. Tab.~\ref{table_performance} shows that our methods outperforms previous methods on RepCount dataset, achieving OBO metric of 0.3267 and MAE metric of 0.4103 compared to the 0.29  and 0.4431 of TransRAC, demonstrating the effectiveness of our proposed method. On UCFRep and QUVA datasets, our model also performs well without any fine-tuning, which indicates the good generalization of our model.

\begin{table*}[h]
\centering\setlength\tabcolsep{5pt}
\caption{Performance of different methods on RepCount test, UCFRep and QUVA when trained on RepCount dataset. The best results are in \textbf{bold} and the second best results are underlined.}
\label{table1}
\begin{tabular}{c|c|c|c|c|c|c}
\hline
\multicolumn{1}{c|}{} & \multicolumn{2}{|c|}{RepCount} & \multicolumn{2}{|c|}{UCFRep} & \multicolumn{2}{|c}{QUVA} \\
\hline
\multicolumn{1}{c|}{Method} & MAE$\downarrow$ & OBO$\uparrow$ & MAE$\downarrow$ & OBO$\uparrow$ & MAE$\downarrow$ & OBO$\uparrow$ \\
\hline
\multicolumn{1}{c|}{RepNet\cite{dwibedi2020counting}} & 0.995 & 0.0134 & 0.9985 & 0.009 & \textbf{0.104} & 0.17 \\
\multicolumn{1}{c|}{Zhang et al.\cite{zhang2020context}} & 0.8786 & 0.1554 & 0.762 & \textbf{0.412} & - & - \\
\multicolumn{1}{c|}{TransRAC\cite{hu2022transrac}} & \underline{0.4431} & \underline{0.2913} & \underline{0.6401} & 0.324 & - & - \\
\multicolumn{1}{c|}{\textbf{Ours}} & \textbf{0.4103} & 
\textbf{0.3267} & \textbf{0.4608} & \underline{0.3333} & 0.4952 & \textbf{0.25} \\
\hline
\end{tabular}
\label{table_performance}
\end{table*}

\begin{figure}[h]
\centering
\includegraphics[width=8cm, height=6cm]{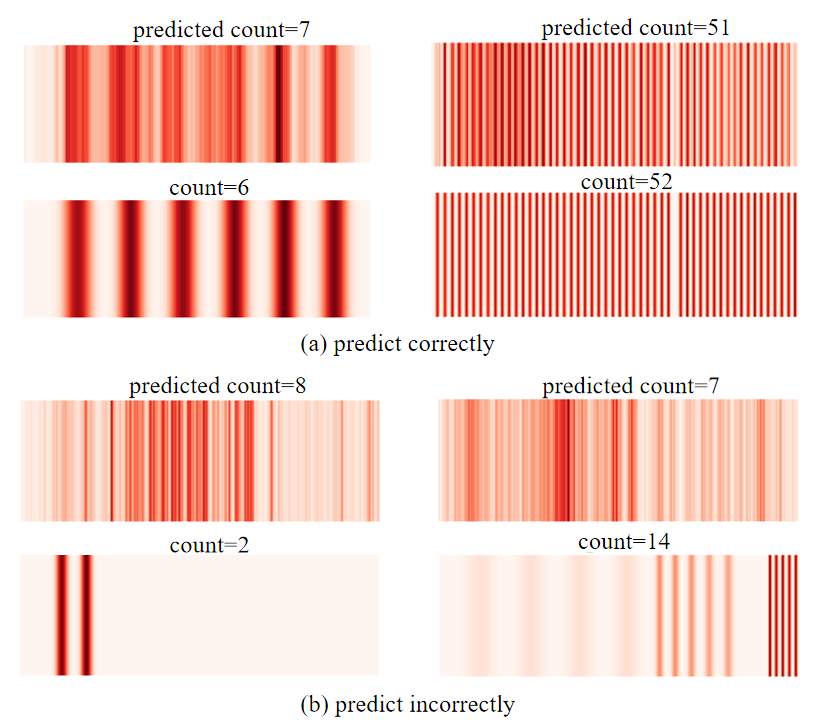}
%
\caption{\label{density map}Visualization of prediction results. The left in (b) indicates that there is interference in the background. The right in (b) shows various duration of actions.} 
\end{figure}

In Figure ~\ref{density map}, we give the visualization of predicted results. As can be seen from the failure cases, our model still have some problems. The left in (b) indicates that interference in the background affects the model's predictions. The right in (b) shows that the duration of actions within the video varies greatly, while our model cannot capture inconsistent action cycles in some extreme cases, which is left to future work.

\subsection{Ablation Study}
We perform several ablations to justify the decisions made while designing the model.

\begin{table}[h]
\centering\setlength\tabcolsep{5pt}
\caption{Ablation study of different frame rate on the RepCount dataset.}
\label{table1}
\begin{tabular}{c|c|c}
\hline
\multicolumn{1}{c|}{frame rate} & \multicolumn{1}{|c|}{MAE$\downarrow$} & \multicolumn{1}{|c}{OBO$\uparrow$} \\
\hline

\multicolumn{1}{c|}{1} & 0.4366 & 0.3 \\

\multicolumn{1}{c|}{2} & 0.6455 & 0.2667 \\

\multicolumn{1}{c|}{3} & 0.5231 & 0.3067 \\

\multicolumn{1}{c|}{4} & 0.4343 & 0.32 \\

\multicolumn{1}{c|}{5 (ours)} & \textbf{0.4103} & \textbf{0.3267} \\

\multicolumn{1}{c|}{6} & 0.4131 & 0.3133 \\
\hline
\end{tabular}
\label{table_framerate}
\end{table}

\begin{table}[h]
\centering\setlength\tabcolsep{5pt}
\label{table1}
\caption{Ablation study of the full temporal resolution and TCNs
on RepCount dataset. The number behind the convolution indicates the kernel size.}
\begin{tabular}{c|c|c|c}
\hline
\multicolumn{1}{c|}{\textbf{TCNs}} & \multicolumn{1}{|c|}{\textbf{Convolution}} & \multicolumn{1}{|c|}{MAE$\downarrow$} & \multicolumn{1}{|c}{OBO$\uparrow$} \\
\hline

\multicolumn{1}{c|}{} & \multicolumn{1}{|c|}{} & 0.4683 & 0.2667 \\

\multicolumn{1}{c|}{\Checkmark} & \multicolumn{1}{|c|}{Vanilla/1} & 0.578 & 0.1933 \\

\multicolumn{1}{c|}{\Checkmark} & \multicolumn{1}{|c|}{Vanilla/3} & 0.4573 & 0.24 \\

\multicolumn{1}{c|}{\Checkmark} & \multicolumn{1}{|c|}{Dilated/3} & \textbf{0.4103} & \textbf{0.3267} \\
\hline

\end{tabular}
\label{table_TCNs}
\end{table}

\begin{table}[h]
\centering\setlength\tabcolsep{5pt}
\label{table1}
\caption{Performance of different number of dilated residual blocks on the RepCount dataset. When there are 6 blocks, both OBO and MAE can achieve the best results}
\begin{tabular}{c|c|c}
\hline
\multicolumn{1}{c|}{\textbf{nums of layer}} & \multicolumn{1}{|c|}{MAE$\downarrow$} & \multicolumn{1}{|c}{OBO$\uparrow$} \\
\hline
\multicolumn{1}{c|}{4} & 0.4126 & 0.3 \\

\multicolumn{1}{c|}{6} & \textbf{0.4103} & \textbf{0.3267} \\

\multicolumn{1}{c|}{8} & 0.4333 & 0.2667 \\

\multicolumn{1}{c|}{10} & 0.486 & 0.2867 \\

\multicolumn{1}{c|}{12} & 0.4172 & 0.26 \\

\multicolumn{1}{c|}{14} & 0.4327 & 0.2667 \\

\multicolumn{1}{c|}{16} & 0.4612 & 0.2733 \\
\hline
\end{tabular}
\label{nums of layers}
\end{table}

\subsubsection{Frame Rate.} In our paper, we are the first to introduce full resolution into repetition counting field, which can provide rich information for network. However, considering the maximum duration of the video, in order to reduce the computational burden, we sample input videos with different frame rates. In Tab. ~\ref{table_framerate}, we compare the performance of different frame rate. Too small frame rate results in redundancy of information, while too large frame rate will lead to ignorance of some repetitions. With the consideration of performance and efficiency, we set the frame rate to 5. 

\subsubsection{Temporal Convolutional Networks (TCNs).} TCNs is a class of time-series models, which is commonly used in action segmentation field. In our model, TCNs contains several dilated residual blocks, which are mainly composed of 1D dilated convolutions. To demonstrate the effect of TCNs, we conduct experiment on the model without TCNs, which means the similarity matrix is directly generated by video features (outputs of video swin transformer). In Tab.~\ref{table_TCNs}, we find that convolution can improve the performance. Under the same conditions, dilated convolution can obtain improvements on MAE by $ 10.28\% $ and OBO by $ 36.13\% $ compared to vanilla convolution, the reason for which may be that dilated convolution has a larger receptive field and can capture similar patterns in long videos.

In Tab. \ref{nums of layers}, we further perform ablations on the number of dilated residual blocks. This observation shows that it is the most appropriate to set the numbers of layers to 6. Too many blocks will cause a decrease in performance.

\section{Conclusion}
In this paper, considering the problems of existing methods in dealing with long videos, we propose a model based on full temporal resolution together with temporal convolutional networks for repetition counting. Our model makes the first attempt to introduce the full temporal resolution into the repetition field. Using dilated convolution can have a huge receptive filed and make it possible to get fined-grained information as well as capturing long-range dependencies. Experimental results show that our model performs better than other video-level models on RepCount dataset and generalizes well on multiple datasets.

\section*{Acknowledgements}
This work is supported in part by the National Natural Science Foundation of China under Grant 62261160652; in part by the National Natural Science Foundation of China under Grant 61733011; in part by the National Natural Science Foundation of China under Grant 62206075; in part by the National Natural Science Foundation of China under Grant 52275013; in part by the GuangDong Basic and Applied Basic Research Foundation under Grant 2021A1515110438; in part by the Guangdong Basic and Applied Basic Research Foundation under Grant 2020B1515120064; in part by the Shenzhen Science and Technology Program under Grant JCYJ20210324120214040; in part by the National Key Research and Development Program of China under Grant 2022YFC3601700.

%
%
\bibliographystyle{splncs04}
\bibliography{reference}

\end{document}